\documentclass[10pt,twocolumn,letterpaper]{article}

\usepackage{iccv}
\usepackage{times}
\usepackage{epsfig}
\usepackage{graphicx}
\usepackage{amsmath}
\usepackage{amssymb}
\usepackage{soul}

\usepackage[pagebackref=true,breaklinks=true,letterpaper=true,colorlinks,bookmarks=false]{hyperref}

\iccvfinalcopy 

\def\iccvPaperID{707} 

\newcommand{\smallsec}[1]{ {\bf #1}}

\newcommand{\myurl}[1]{\texttt{#1}}

\ificcvfinal\fi
\begin{document}

\title{What Actions are Needed for Understanding Human Actions in Videos?}

\author{Gunnar A. Sigurdsson \ \ \ \ 
Olga Russakovsky \ \ \ 
Abhinav Gupta \\
Carnegie Mellon University \\ 
\myurl{github.com/gsig/actions-for-actions}
}

\maketitle

\begin{abstract}

What is the right way to reason about human activities? What directions forward are most promising?
In this work, we analyze the current state of human activity understanding in videos. The goal of this paper is to examine datasets, evaluation metrics, algorithms, and potential future directions. We look at the qualitative attributes that define activities such as pose variability, brevity, and density. The experiments consider multiple state-of-the-art algorithms and multiple datasets. The results demonstrate that while there is inherent ambiguity in the temporal extent of activities, current datasets still permit effective benchmarking. We discover that fine-grained understanding of objects and pose when combined with temporal reasoning is likely to yield substantial improvements in algorithmic accuracy. We present the many kinds of information that will be needed to achieve substantial gains in activity understanding: objects, verbs, intent, and sequential reasoning. The software and additional information will be made available to provide other researchers detailed diagnostics to understand their own algorithms.

\end{abstract}

\begin{figure*}
\includegraphics[width=1.0\linewidth]{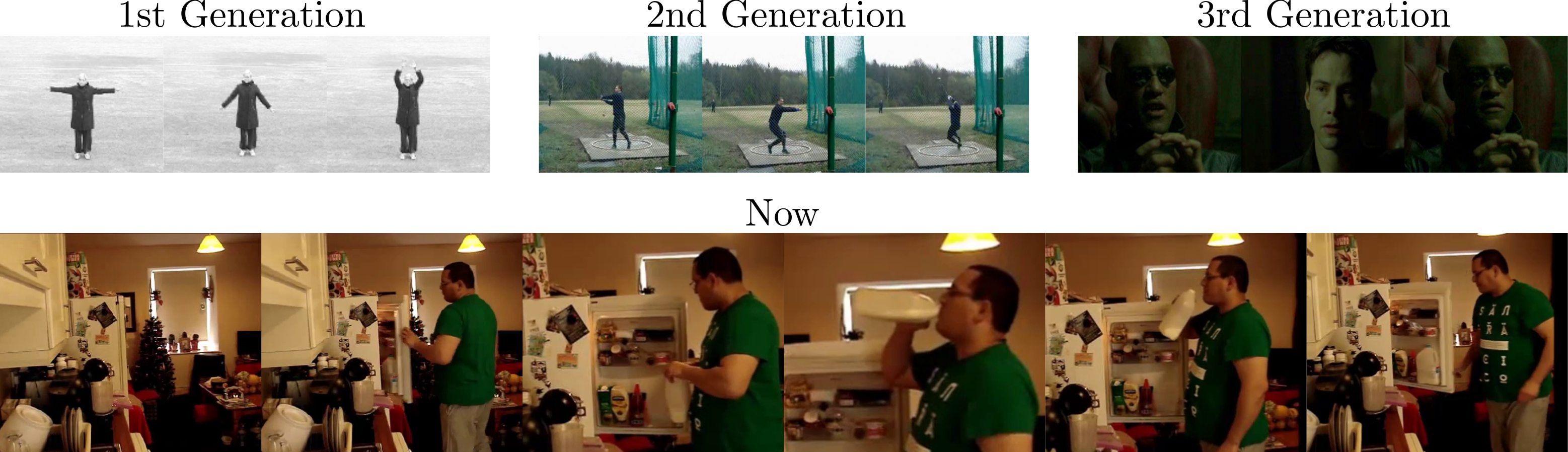}
\caption{Now that the field of activity recognition has moved on from simple motions (KTH~\cite{schuldt2004recognizing}), sports (UCF101~\cite{UCF101}), and isolated activities in movies (HMDB51~\cite{kuehne2011hmdb}) to cluttered sequences of home activities (Charades~\cite{sigurdsson2016hollywood}), how should we think about activities? What are the categories? Do activities have well-defined boundaries? In the highlighted video from Charades, a person ``walks to the kitchen,'' ``opens the fridge,'' ``grabs some milk,'' ``opens the bottle,'' ``drinks from the bottle,'' ``puts it back,'' and ``closes the fridge.'' In this work, we discuss the problem of how to think about when, where, and what the person is doing at any given time.}
\label{fig:teaser}
\end{figure*}

\section{Introduction}
Over the last few years, there has been significant advances in the field of static image understanding. There is absolutely no doubt that we are now closer to solving tasks such as image classification, object detection, and even semantic segmentation. On the other hand, when it comes to video understanding we are still struggling to figure out basic questions such as: What is an activity and how should we represent it? Do activities have well-defined spatial and temporal extent? What role do goals and intentions play in defining and understanding activities?

A significant problem in the past has been the absence of good datasets for activity detection and recognition. Most of the major advances in the field of object recognition have come with the creation of generic datasets such as PASCAL~\cite{Everingham15}, ImageNet~\cite{Deng09} and COCO~\cite{lin2014microsoft}. These datasets  helped define the problem scope and the evaluation metrics, as well as revealed the shortcomings of existing approaches.

On the other hand, historically video datasets such as UCF101~\cite{UCF101} have been biased and have corresponded to activities that are hardly seen in daily lives. However, in recent years, things have started to change with the arrival of large-scale video datasets depicting a variety of complex human activities in untrimmed videos. Datasets such as ActivityNet~\cite{caba2015activitynet}, Sports1M~\cite{karpathy2014large}, Charades~\cite{sigurdsson2016hollywood}, and MultiTHUMOS~\cite{THUMOS15,yeung2015every} have revitalized activity recognition and inspired new research and new ideas.

But before we move forward and define the benchmarks, we believe it is worth pausing and thoroughly analyzing this novel domain. What does the data show about the right categories for recognition in case of activities? Do existing approaches scale with increasing complexity of activities categories, video data, or temporal relationships between activities? Are the hypothesized new avenues of studying context, objects, or intentions worthwhile: Do these really help in understanding videos?

This paper provides an in-depth analysis of the new generation of video datasets, human annotators, activity categories, recognition approaches, and above all possible new cues for video understanding. Concretely, we examine:

   \smallsec{What are the right questions to ask?} Here we investigate some fundamental questions regarding the problem of activity recognition. First, we ask what are the right categories for activities? Unlike objects where semantic categories are  somewhat well-defined, activity categories defined by verbs are relatively few in number. Should we model one activity category called ``open'' or should we model ``open suitcase'', ``open windows'' and ``open curtain'' by different categories. We also ask a fundamental question: whether we should perform classification or localization. Do humans agree on temporal activity boundaries? And if they do not, is it worth exploring localization at all?
    
    \smallsec{What are existing approaches learning, and are those the right things?} In the next set of analyses, we explore the current algorithms for activity classification and localization. We define attributes for quantifying video and category complexity and evaluate algorithms with respect to these attributes. Do current approaches work better when there are multiple activities per video by exploiting the context? Does large variation in pose among categories help in activity classification? The analyses presented in this section have been combined into a single open-sourced tool that automatically renders a summary of the diagnostics and suggest improvements to existing algorithms. 
    
    \smallsec{What directions seem most promising?} By considering various ideal components of activity understanding, we evaluate how can we get the next big gain. Should we explore intentions, verbs, and sequences of activities, or only focus on jointly reasoning over objects and activities?
    

\section{Evolution of Activity Recognition}
\label{sec:evolution}

Starting with the KTH action dataset~\cite{schuldt2004recognizing} we have observed significant advances in understanding activities through new datasets and algorithms. KTH combined with the idea that activities are \emph{motions} sparked numerous advances~\cite{STIP05,HOG3D,HOF,kovashka2010learning,MBH06}. With increasingly complex datasets such as UCF101~\cite{UCF101} and others~\cite{karpathy2014large,kuehne2011hmdb,liu2009recognizing,THUMOS15} came new challenges, including scale, background clutter, and action complexity, in turn leading to improved algorithms~\cite{WangIDT13,simonyan2014twostream,packer2012combined,3DCNN}. However, these datasets still focused on short individual clips of activities, and commonly sports, which encouraged the next wave of datasets that focus on sequences of everyday activities~\cite{pirsiavash2012detecting,caba2015activitynet,yeung2015every,sigurdsson2016hollywood}.

These recent datasets present different challenges for activity understanding. ActivityNet~\cite{caba2015activitynet} includes many activity categories and a dense hierarchy of activities, although each video is only associated with one activity label. THUMOS~\cite{THUMOS15} and its extension MultiTHUMOS~\cite{yeung2015every} provide multi-label sports videos. Charades~\cite{sigurdsson2016hollywood} contains diverse and densely annotated videos of common daily activities occurring in a variety of scenes. 

\smallsec{Our setup:} In this work, much of our evaluation focuses on understanding the scope of and the interaction between different activities. Thus we choose Charades~\cite{sigurdsson2016hollywood} as the best testbed for our analysis, and introduce MultiTHUMOS~\cite{yeung2015every}, THUMOS~\cite{THUMOS15}, and ActivityNet~\cite{caba2015activitynet} as needed to establish the generality of our conclusion. All datasets use the same normalized mAP metric that is robust to different ratios of positives to negatives as well as different number of categories~\cite{hoiem2012diagnosing}. Charades contains $9{,}848$ videos, split into $7{,}985$ for training and $1{,}863$ for testing. It contains annotations of 157 activity categories such as ``drinking from a cup'' and ``watching the tv'' which happen naturally along with other categories. ``watching the tv'' might for example occur with ``lying on the couch'' and ``snuggling with a blanket'', or ``drinking from a cup''. 
That is, activity categories have moved from capturing \emph{verbs} to capturing a variety of \emph{(object,verb)} pairs, and we begin our investigation by analyzing this distinction in more detail.

\smallsec{Baselines:} We evaluate Two-Stream~\cite{simonyan2014twostream}\footnote{The Two-Stream network uses a soft-max loss function and randomly samples positive training examples from each class at training time.}, Improved Dense Trajectories (IDT)~\cite{WangIDT13}, LSTM on top of VGG-16~\cite{cnnlstm}, and two recent approaches: ActionVLAD~\cite{Girdhar_17a_ActionVLAD} with sophisticated spatio-temporal pooling and Asynchronous Temporal Fields~\cite{sigurdsson2017asynchronous} with a deep structured model.\footnote{We primarily use the test set predictions of these models released by~\cite{sigurdsson2017asynchronous} available at \myurl{github.com/gsig/temporal-fields}; for ActionVLAD we use predictions provided by~\cite{Girdhar_17a_ActionVLAD}. }

\section{What are the right questions to ask?}

To start our discussion about activities, let us establish what we want to learn. 
When we talk about activities, we are referring to anything a person is doing, regardless of whether the person is intentionally and actively altering the environment, or simply sitting still. In this section, we will first look at how to define activity categories, and then investigate the temporal extents of activities.

\subsection{What are the right activity categories?}
\label{sec:right_questions}
Should we focus our analysis on general categories such as ``drinking'', or more specific, such as ``drinking from cup in the living room''?
Verbs such as ``drinking'' and ''running'' are unique on their own, but verbs such as ``take'' and ``put'' are ambiguous unless nouns and even prepositions are included: ''take medication'', ``take shoes'', ``take off shoes''. That is, nouns and verbs form atomic units of actions.

To verify this property of verbs, we ran a human study using workers on Amazon Mechanical Turk. We presented them with Charades videos~\cite{sigurdsson2016hollywood}, and asked workers to select which of the 157 activities are present in the video. We looked at the likelihood a worker will choose an activity B when activity A is present in the video but not B. 
We found that people considered verbs to be relatively more ambiguous. That is, given the verb, workers confused activities with different objects, e.g., ``holding a cup'' vs ``holding a broom,'' only $0.3\%$ of the time. However given the object, there was more confusion among different verbs, e.g., ``holding a cup'' vs ``drinking from a cup,'' $1.3\%$ of the time. A similar pattern of confusion between similar categories is seen by state-of-the-art algorithms in Sec.~\ref{sec:errors}.

To quantitatively determine the distinctiveness of the different types of categories, we looked at pose similarity (Sec.~\ref{sec:person}) across categories on the Charades dataset, both for verb supercategories, and (verb,noun) categories. As expected, even with this simple metric there was more distinctiveness across categories for (noun,verb) than verbs ($p{<}0.01$). 
This reinforces that verbs alone lack the clarity to provide clear supervision for learning.

\subsection{Do Activities Have Temporal Extents?}
\label{sec:temporal_extents}

Physical objects have clear boundaries defined by their physical extent, which is visible as a depth change with respect to the viewer. Activities, however, have few such clear boundaries. How should we evaluate \emph{when} each activity is happening? Would two people even agree on this?

We looked at how well human annotators agreed with the activity boundaries by asking Amazon Mechanical Turk workers to re-annotate the extent of actions in the Charades and MultiTHUMOS videos. We make three observations.

First, the average agreement with ground truth was only $72.5\%$ intersection-over-union (IOU)\footnote{Equivalently, only $76.1\%$ were over $50\%$ IOU.} for Charades and $58.7\%$ IOU in MultiTHUMOS, indicating that temporal extent of an activity is ambiguous even for humans. The median starting error was less than the ending error ($0.9{\pm}0.8$ sec compared to $1.4{\pm}1.4$ sec in Charades), which suggests that more of the confusion is about the end of the activity. 

Second, there is a significant positive correlation between IOU agreement and the duration of activity ($\rho{=}0.50$) suggesting that longer activities tend to be easier for humans to localize. Further, humans tend to better agree on the starting point of longer activities compared to shorter activities: the starting error in seconds decreases with longer duration ($\rho{=}0.18$). This suggests that categories of temporally brief activities may require more careful annotation. 

Finally, the difference in IOU agreement between categories in Charades was lower than average IOU agreement ($13.0\%$ IOU standard deviation compared to $27.5\%$ average IOU), implying that the ambiguity in temporal boundaries is a common problem for many activity categories. 

This analysis suggests that there is inherent ambiguity in precisely localizing activities, which primarily depends on the length of the activity, and evaluation metrics must account for this. Furthermore, this suggests that algorithms might benefit from treating the activity boundaries as fluid, which is related to jointly reasoning about the whole video and the boundaries therein.

\smallsec{Can we evaluate temporal localization?} This raises a natural question: if there is such inherent ambiguity in localizing activities, is evaluating localization simply too erroneous? To analyze this, we experimented with omitting ambiguous boundary regions from the test set. Concretely, for a ground truth action instance with temporal extent $[t_{1}, t_{2}]$, we define the boundary as $B{=}[t_{1}{-}\alpha, t_1{+}\alpha] \cup [t_{2}{-}\alpha, t_2{+}\alpha]$ with $\alpha{=}(t_{2}{-}t_{1})/3$.

We again consider human annotations and compute consensus IOU ignoring the boundary region in both the intersection and the union. The human consensus increases from $72.5\%$ IOU to $79.8\%$ IOU. 
We found that the center ($33\%$) of an activity was likely to be agreed upon, where $82.7\%$ of the center was covered by a subsequent annotator.\footnote{Training a Two-Stream (RGB)~\cite{simonyan2014twostream} baseline on only this part of the activity yielded an improvement in classification, $15.9\%$ to $16.7\%$ mAP.}

To investigate if this is significant on algorithmic evaluation, we use a Two-Stream network~\cite{simonyan2014twostream} (Sec.~\ref{sec:evolution}), which outputs per-frame predictions for 157 activities. We evaluate its per-frame accuracy following the Charades metric~\cite{sigurdsson2016hollywood} ignoring the ambiguous boundary regions. Accuracy increases from $9.6\%$ mAP to $10.9\%$ mAP  (0.1-4.3\% increase on other datasets). Looking at state-of-the-art baselines on Charades (Sec.~\ref{sec:evolution}) we find that they increase by $0.96 \pm 0.33 \%$ mAP where $0.33\%$ mAP is less than the average difference between methods $1.26\%$ mAP. Thus, this does not affect order for evaluation.
This suggests that despite boundary ambiguity, current datasets allow us to understand, learn from, and evaluate the temporal extents of activities.


\smallsec{Should we evaluate temporal localization?} We ask one final question. When videos are short, is it perhaps unnecessary to worry about localization at all?  We measured how well a perfect video classifier emitting the same binary prediction for every frame would fare on localization accuracy. In Charades, videos are $30.1$ seconds on average and localization accuracy of this oracle was $56.9\%$ mAP. That is, a perfect classifier would automatically do 5 times better than current state-of-the-art~\cite{sigurdsson2017asynchronous} on activity localization. Similarly, a perfect classifier improves over our localization baseline on all the datasets. 

This suggests that focusing our attention on gaining more insight into activity classification would naturally yield significant improvements in localization accuracy as well. Further, as we will see, understanding temporal relationships is important for perfect classification; with the complete understanding of activities and their temporal relationships needed to get perfect classification accuracy, we stand to benefit in terms of localization as well. In the rest of this paper, we focus on classification accuracy, but report localization accuracy to gain a deeper understanding.

\section{What are existing approaches learning?}
\label{sec:category}

Having concluded that (1) we should be reasoning about activities as \emph{(verb,object)} pairs rather than just \emph{verb}, that (2) temporal boundaries of activities are ambiguous but nevertheless meaningful, and that (3) classification of short videos is a reasonable proxy for temporal localization, we now dig deeply into the state of modern activity recognition. What are our algorithms learning? In Sec.~\ref{sec:errors} we start by analyzing the errors made by state-of-the-art approaches; in Sec.~\ref{sec:trainingdata} we analyze the effect of training data and activity categories; in Sec.~\ref{sec:temporalreasoning} we look at the importance of temporal reasoning; and in Sec.~\ref{sec:person} we emphasize the importance of person-based reasoning. All annotated attributes along with software to analyze any new algorithm along with diagnostics from the software for various algorithms are available at \myurl{github.com/gsig/actions-for-actions}.

\smallsec{Setup:}
The models described in Sec.~\ref{sec:evolution} are trained on the Charades training videos and we evaluate classification mAP on the Charades test set unless stated otherwise.
Accuracy is measured using mAP, which is normalized~\cite{hoiem2012diagnosing} as to be comparable across different subsets of the data with different numbers of positive and negative examples. Please refer to the Appendix for details.

The category plots are generated from $(x,y)$ pairs where $x$ is a an attribute for a category and $y$ is the classification performance for the category. Finally, the pairs are clustered by the $x$ coordinate and the average of the $y$ coordinates visualized. Error bars for category plots represent one standard deviation around Two-Stream based on the $y$ values in each cluster. In this section we report Pearson's $\rho$ correlation between $x$ and $y$. The video plots are generated similarly by clustering the videos based on the attributes. Finally the mAP is calculated in that group of videos. Error bars for video plots represent the $95\%$ confidence interval around Two-Stream obtained via bootstapping~\cite{everingham2013assessing}.

\subsection{Analyzing correct and incorrect detections}
\label{sec:errors}
\noindent \emph{What kind of mistakes are made by current methods?}

\begin{figure}[tb]
\includegraphics[width=1.0\linewidth]{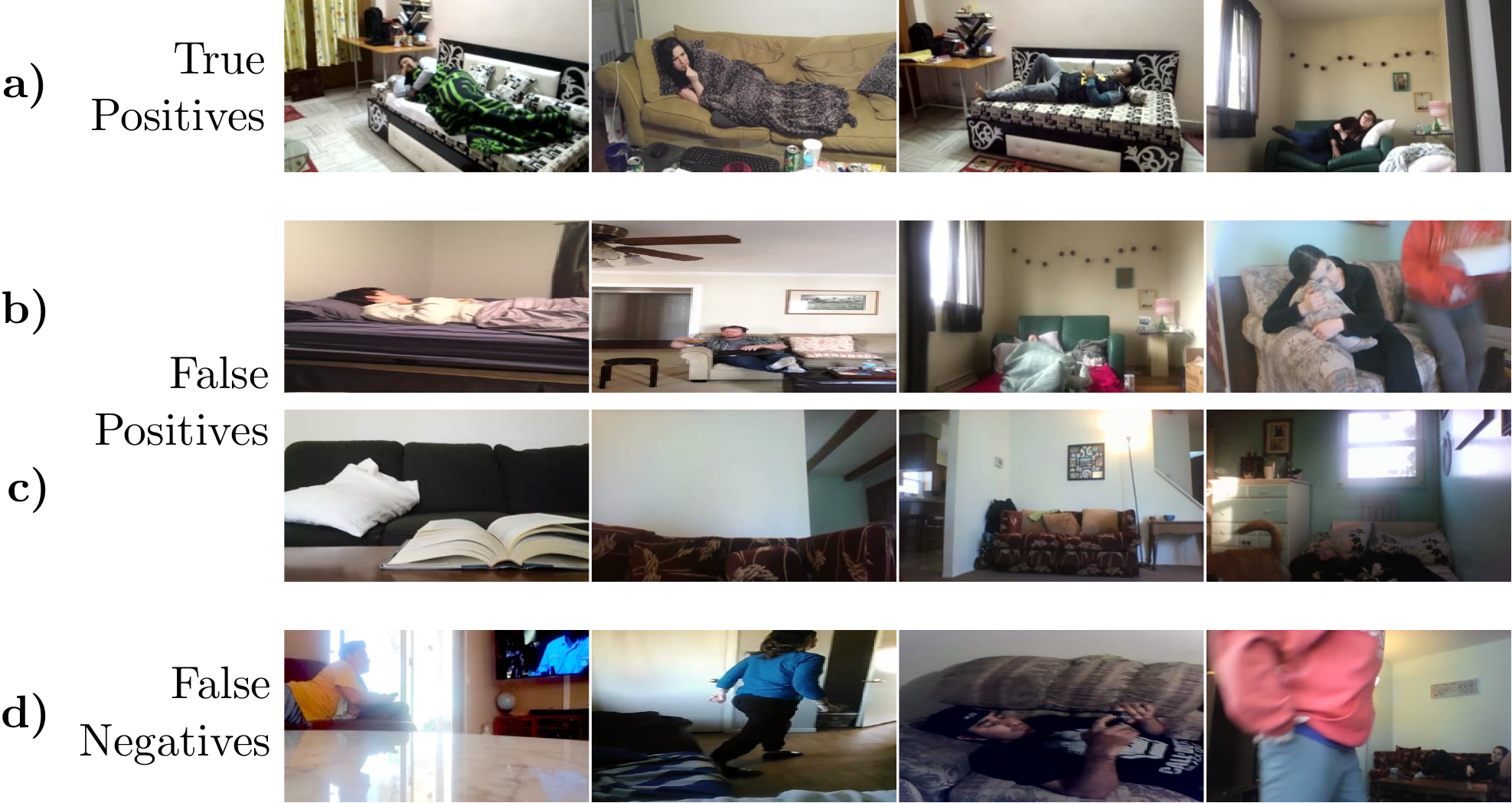}
\caption{
Example results from a Two-stream network~\cite{simonyan2014twostream} on per-frame classification of the ``Lying on a couch'' action in Charades. a) High-scoring true positives commonly include canonical views of both a couch and a person. Top false positives often include b) confusion with other objects (e.g., beds) or verbs (e.g., sitting) and c) the correct scene but with an absent action (e.g., vacant couches). d) Top false negatives include unusual viewpoints.}
\label{fig:visual}
\end{figure}

To understand what current methods are learning and motivate the rest of this section, we start by highlighting some of the errors made by current methods.
First, we look at visual examples of the errors that a Two-Stream Network~\cite{simonyan2014twostream} makes on Charades. In Fig.~\ref{fig:visual} we see correct classifications, as well as three types of errors. The figure suggests that
1) models need to learn how to reason about similar categories (Sec.~\ref{sec:trainingdata});
2) methods have to develop temporal understanding that can suppress temporally similar but semantically different information (Sec.~\ref{sec:temporalreasoning}); and,
3) models need to learn about humans and not assume if couch is detected, then ``Lying on a couch'' is present (Sec.~\ref{sec:person}).

\begin{figure}[tb]
\includegraphics[width=1.0\linewidth]{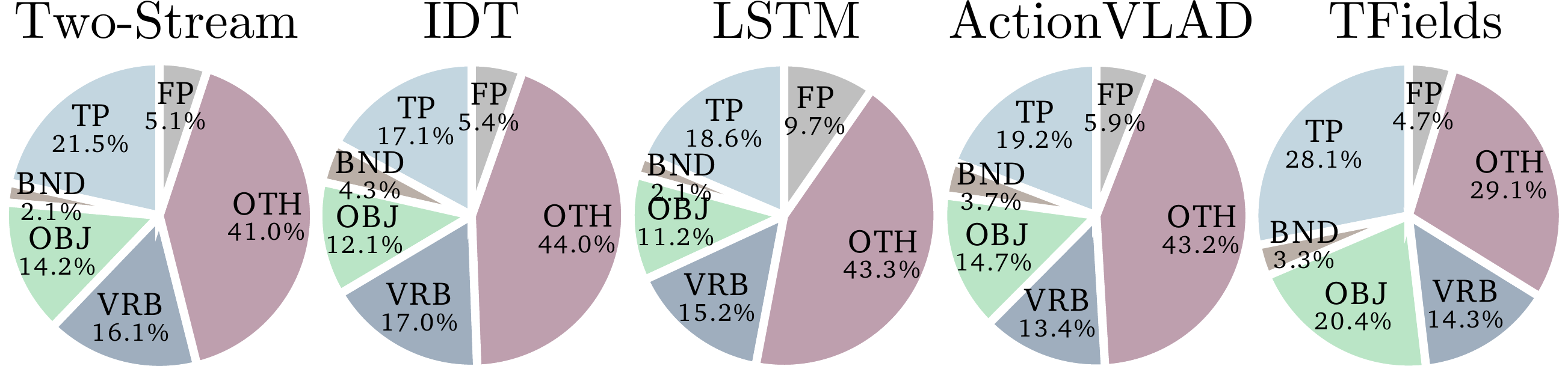}
\caption{Fraction of top ranked predictions for each class that are correct (TP), on the boundary (BND), other class with same object (OBJ), other class with same verb (VRB), other class with neither (OTH), or no class (FP). Inspired by Hoiem et al.~\cite{hoiem2012diagnosing}.}
\label{fig:pies}
\end{figure}

\begin{figure}[tb]
\includegraphics[width=1.0\linewidth]{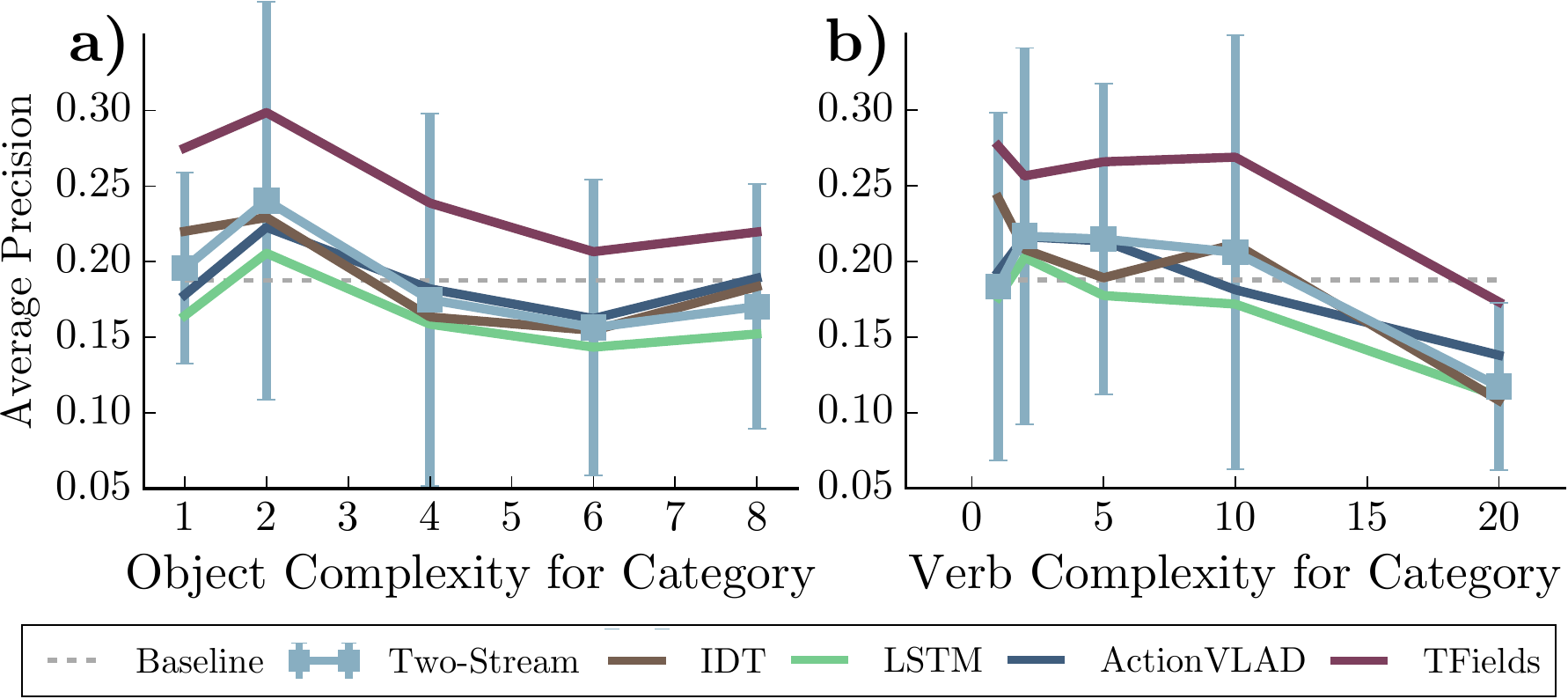}
\caption{Accuracy of activity categories for multiple methods as a function of how many categories share its object/verb.} 
\label{fig:objverb}
\end{figure}

To provide further insight into the algorithms we provide brief overview of the types of errors made by multiple algorithms. In Fig.~\ref{fig:pies} we look at the relative types of errors. First, we note that not many errors are made close to the boundary. However, we can see that significant confusion is among similar categories, both for verbs and objects, interestingly TFields~\cite{sigurdsson2017asynchronous} confuse unrelated categories much less ($29.1\%$ compared to $41.0-44.0\%$), but at the cost of confusing categories with similar object. 
In fact, for a given category the more categories share the object or verb, the worse the accuracy. This is visualized in Fig.~\ref{fig:objverb}, where we consider how many categories share an object/verb with the given category (\emph{Object/Verb complexity}). This suggests that moving forward fine-grained discrimination between activities with similar objects and verbs is needed.

\subsection{Training Data}
\label{sec:trainingdata}
\noindent \emph{Are current methods limited by the available data? How do we handle imbalanced data and similar categories?}

To understand how training data affects current algorithms, we first look at the relationship between amount of data and accuracy to understand how to better make use of data and what kind of data we need more of.

We train a Two-Stream (RGB)~\cite{simonyan2014twostream} baseline and focus on Charades, since it has naturally occurring long-tailed distribution of classes. With all of the available training data we get $15.6\%$ mAP, and with $\frac{7}{10}$ and $\frac{3}{10}$ of the data we get $14.7\%$ and $11.6\%$ mAP. This is expected---more data is better. 
However, what kind of more data is better? For example there is not a clear relationship for larger categories to have higher accuracy in the datasets ($\rho{<}0.10$). Why is more data not helping these categories?

\smallsec{Focus on small categories.} 
First, we note that most categories seem to benefit from more data even in other categories. Since Charades has different number examples in each class, removing $\frac{1}{2}$ of the data removes many more examples from large categories than small. Even so, small categories see a larger decrease in accuracy ($\rho{=}0.18$, $p{=}0.03$). In the extreme, limiting all categories to at most 100 examples, there is still no significant relationship between category size and drop in accuracy.

The only significant correlation we found with this drop in accuracy when limiting all categories to at most 100 examples was with the number of similar categories (that share the same object/verb). Categories with more similar categories had more reduction in performance ($\rho{=}0.18$, $p{<}0.05$). This suggests that \emph{there is no such thing as balanced data}. Any attempts at reweighting and resampling without considering similarity between categories is unlikely to help all small categories uniformly. For example, when limiting categories to at most 100 samples, the range in relative change in accuracy is from a $65.8\%$ \emph{decrease} (``holding a laptop'') to $52.2\%$ \emph{increase} (``standing on a chair''). The highest drop is in a category that has 5 categories that share the object, and 18 categories that share the verb. ``Standing on a chair'' however is a relatively unique activity with only 42 examples. Investigating this large difference in performance when balancing the data will be important to fully harness this kind of activity data.

\smallsec{Focus on large categories.} 
Categories that are naturally ubiquitous and have many examples in naturally long-tailed datasets seem to have additional complexity that outweighs the advantage of having more training data.
For example in Charades, they have slightly more pose diversity: concretely, the correlation between the number of training examples of a category and its pose diversity (as defined in Sec.~\ref{sec:person}) is $\rho{=}0.09$. Second, they tend to have less inter-class variation. For example, categories with more examples have more categories that share the same object $\rho{=}0.13$ (``object complexity'' in Sec.~\ref{sec:errors}). Thus more common actions may in fact be more challenging to learn. 

\begin{figure}[tb]
\includegraphics[width=1.0\linewidth]{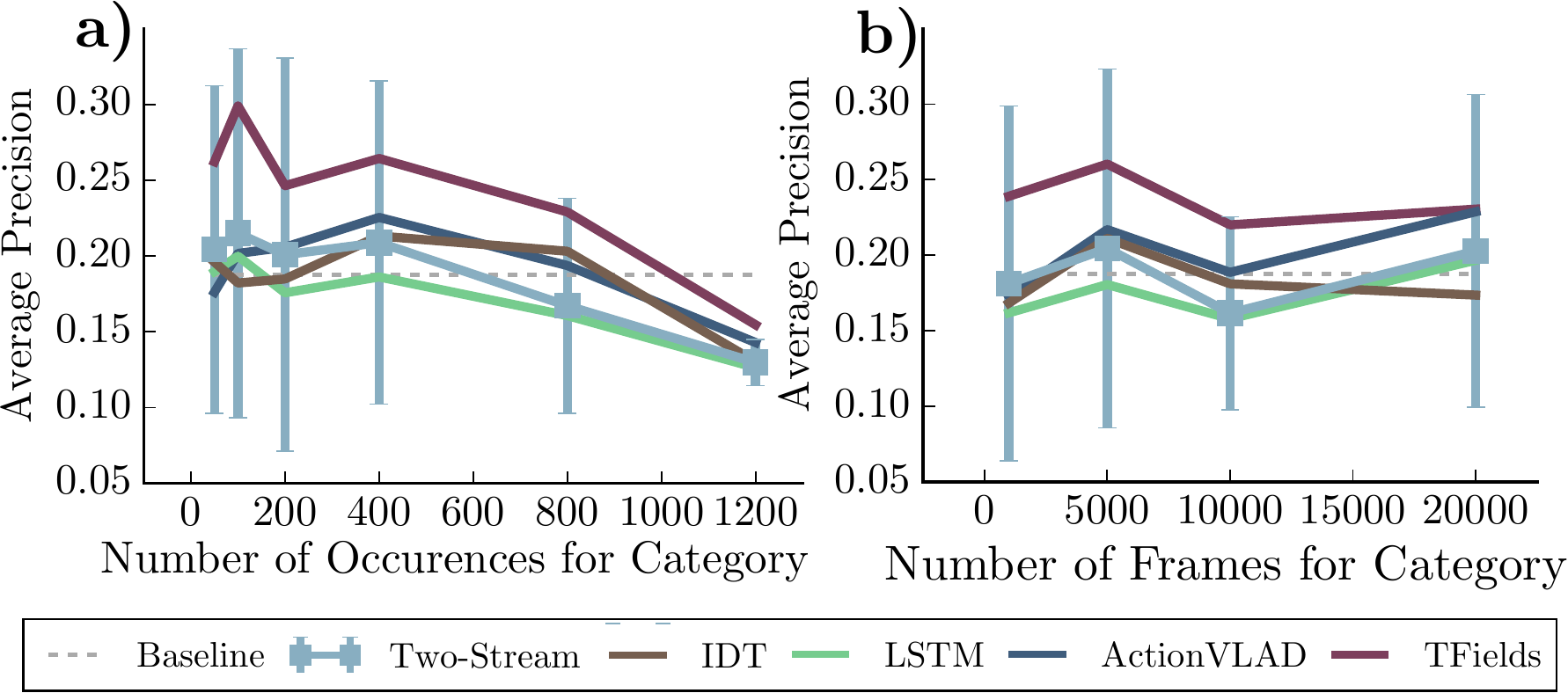}
\caption{Accuracy of activity categories for multiple methods as a function of training examples/frames.} 
\label{fig:numexamples}
\end{figure}

Interestingly, looking at current methods, we find that the main improvement in accuracy does not come from models that are better able to make use of the wealth of data in large categories, but rather in small categories. This is visualized in Fig.~\ref{fig:numexamples} both for number of training examples as well as training frames. This both suggests that developing models that generalize well across categories is clearly beneficial, but also that more expressive models are needed for large categories, perhaps by dividing them into subcategories.

\subsection{Temporal Reasoning}
\label{sec:temporalreasoning}
\noindent \emph{How does the temporal extent of activities affect accuracy? Should we think about activities in terms of high-level temporal reasoning?}

Given that Two-Stream networks, which operate on isolated RGB and Optical Flow frames, are on par with state-of-the-art on many recent datasets, it is either that temporal reasoning beyond instantaneous motion is not needed for activity recognition, or current methods are missing important pieces of temporal reasoning. In this section we look at current algorithms at increasing granularity of temporal reasoning: motion, continuity, and temporal context.

\begin{figure}[tb]
\includegraphics[width=1.0\linewidth]{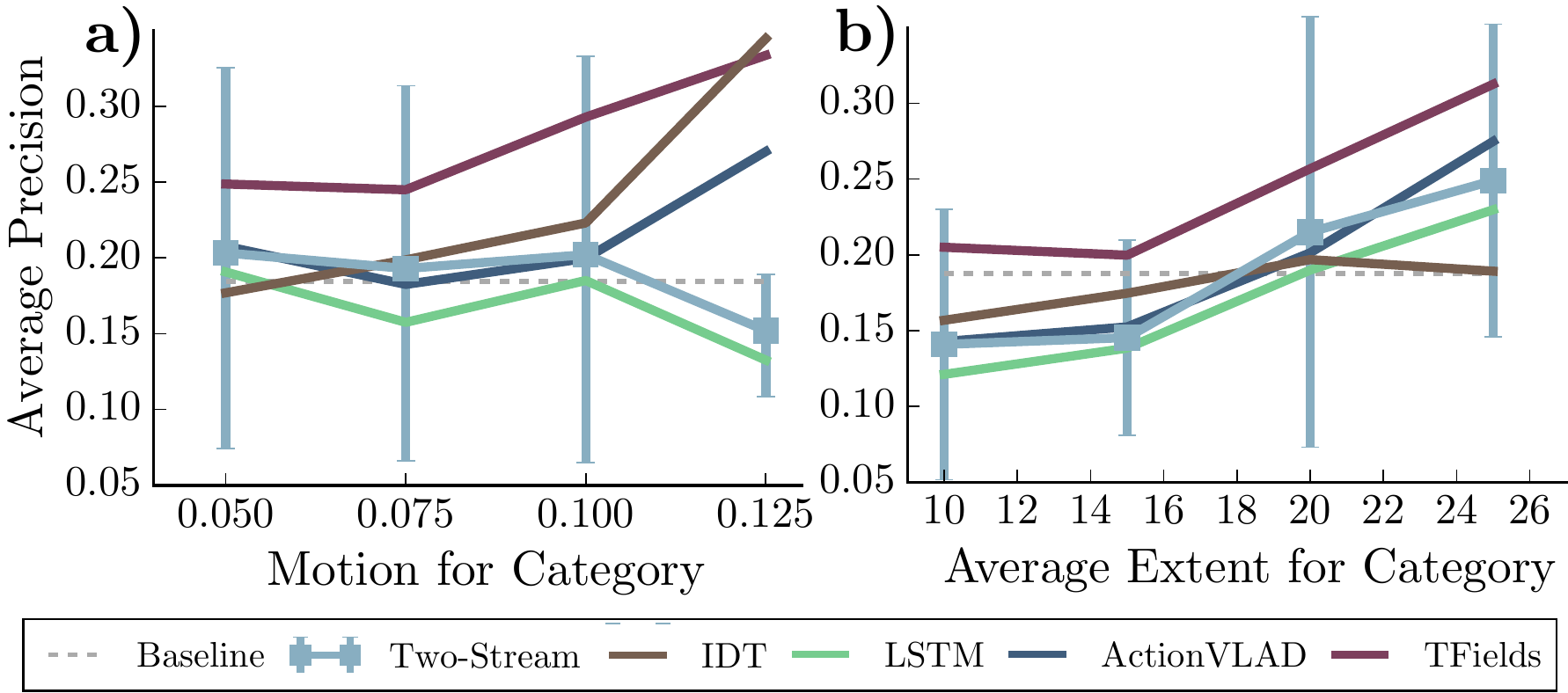}
\caption{Accuracy as a function of motion and temporal extent for categories in Charades.} 
\label{fig:motion}
\end{figure}

\smallsec{Motion and temporal continuity.}
Most activities involve movement that causes blurred frames, intermittent observability, and large visual variation. In theory, algorithmic approaches must be robust to these effects by combining multiple predictions over time.
To analyze how well current algorithms combine predictions to, for example, reason in the presence of motion, we consider average amount of optical flow in a given category on average (\emph{Motion for Category}) and the average temporal extent of activities in each category (\emph{Average Extent for Category}). This is visualized in Fig.~\ref{fig:motion}.
We find that instantaneous motion affects algorithms differently, for example, more temporal reasoning seems to help (IDT~\cite{WangIDT13}, TFields~\cite{sigurdsson2017asynchronous}) algorithms be robust to motion. Furthermore, short actions do noticeably worse on the datasets ($\rho{=}0.21{-}0.42$).
We could expect that longer activities indeed do better because they have more training data, but this view was refuted in Sec.~\ref{sec:trainingdata}. This suggests that current methods are better at modeling longer activities than shorter ones, implying that more emphasis may be needed on understanding shorter activities. 
This both suggests that brief patterns, motion and short actions, need more temporal reasoning to be understood.
One potential avenue for exploration would be combining the trajectory representation in IDT (which appears to help on shorter activities) with the benefits of longer-term pooling of Two-Stream (which works well on longer activities). 

How well do these algorithms combine predictions over time? As it turns out, naive temporal smoothing of the predictions helps improve localization and classification accuracy, where all methods except LSTM benefit from averaging predictions over a window $4\%$ of the video size ($1.2$ sec on average). However, for optimally smoothed Two-Stream for example, the relative change in accuracy for individual classes varies from $24.7\%$ decrease (``Holding a broom'') to $32.4\%$ increase (``Closing a window''). Furthermore, smoothing helps larger classes more ($\rho{=}0.22$, $p{<}0.01$), but does not help classes with much motion. This leads to the conclusion that combining predictions over time is a non-trivial problem that must be addressed in future work.

\smallsec{Temporal Context.}
Moving to larger temporal scales, we now investigate how current methods utilize temporal context. 
We noticed that for videos with 1-4 activities per video, Two-Stream~\cite{simonyan2014twostream} obtains $22.1\%$ mAP, for videos with 14 or more, the accuracy is $16.6\%$ mAP. One could expect that with more activities, the additional context could be used to improve performance, but the additional complexity seems to outweigh any additional context. This pattern also applies to temporal overlap (multiple activities happening at a given instant) where classes that overlap frequently in the training data are significantly more confused at test time ($\rho{=}0.28$ correlation between percent of training frames where classes overlap, versus score given to class A when class B is present but not A).

\begin{figure}[tb]
\centering
\includegraphics[width=0.6\linewidth]{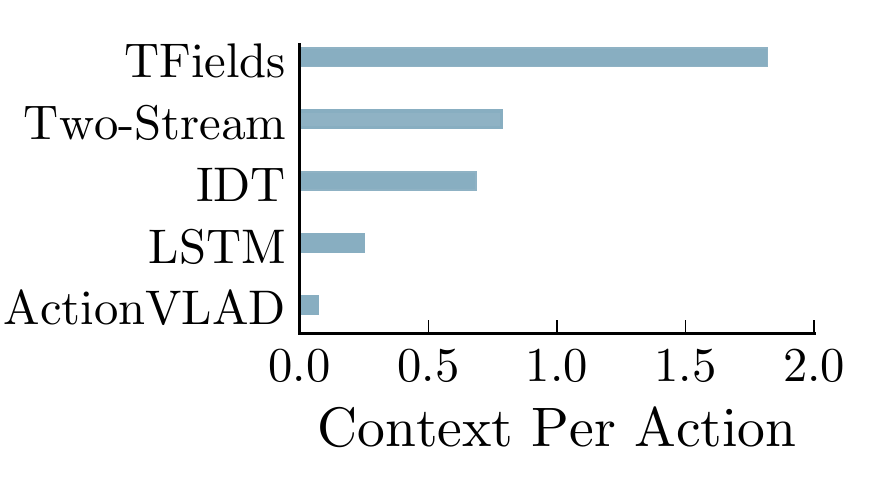}
\caption{How many actions positively influence each action by being present. Presented for multiple algorithms on Charades.} 
\label{fig:context}
\end{figure}

In fact, the methods vary significantly in terms of how much they use context. In Fig.~\ref{fig:context} we visualize how much each method benefits from context; we measure for each action how many other actions on average increase the prediction confidence of that action by being present in a video. We consider video classification, and since all methods do combine their predictions to make a video prediction, context is being used in some form by all methods. We observe that high-level temporal modeling in TFields~\cite{sigurdsson2017asynchronous} helps utilize context, and is important moving forward. 

\subsection{Person-based Reasoning}
\label{sec:person}
\noindent \emph{Should activity recognition be image-based or person-based? Should activity recognition models be explicitly taught what a person is?}

In CNNs for object recognition, it has been observed that object parts are discovered automatically~\cite{zhou2014object}. Ideally, we would want the person to be automatically discovered by activity recognition systems. But is this happening now?

\begin{figure}[tb]
\includegraphics[width=1.0\linewidth]{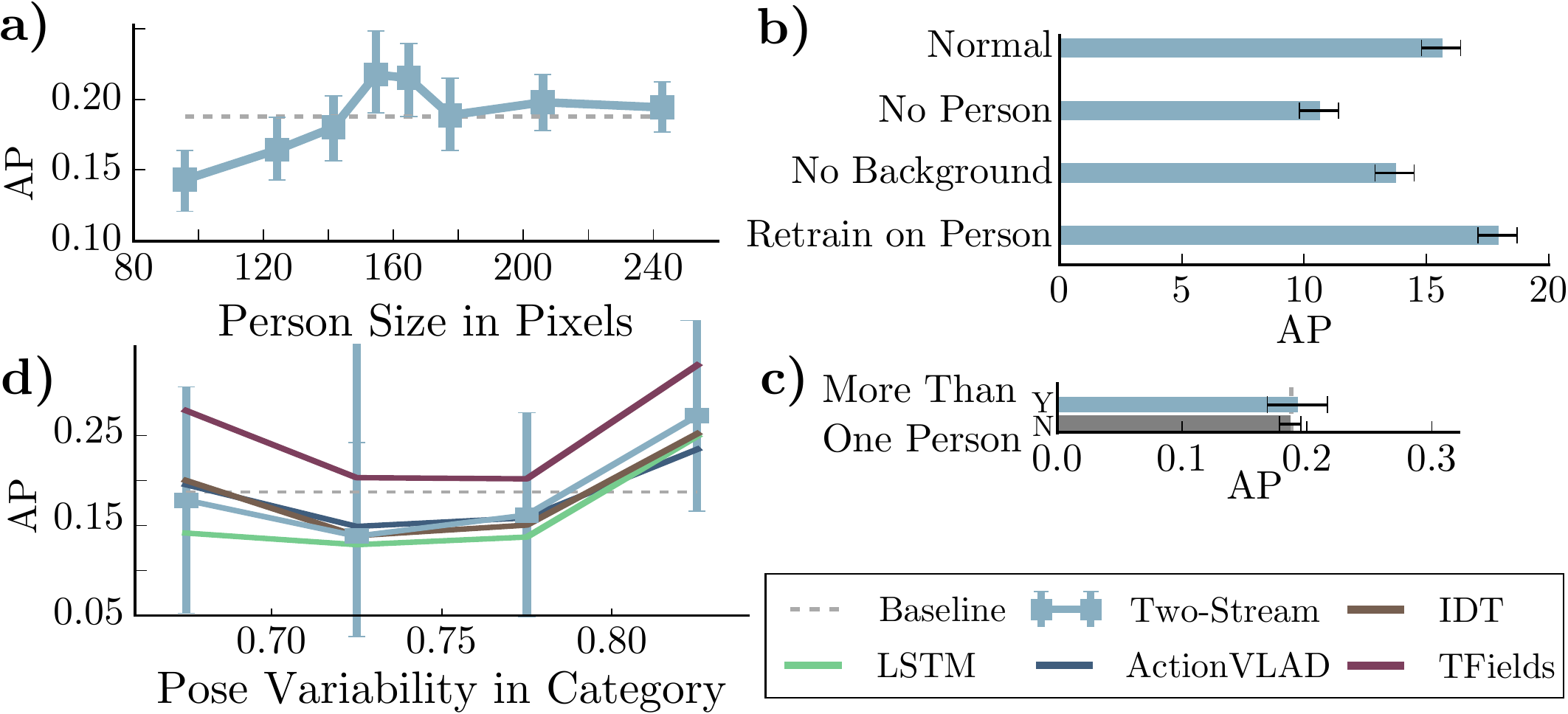}
\caption{Supporting plots for Sec.~\ref{sec:person}. Visualizing the impact of attributes on classification performance in the Charades Dataset.}
\label{fig:person}
\end{figure}

\smallsec{Person location in the frame.}
First, we look at the average size of a person in a video as measured by the highest-scoring per-frame Faster-RCNN bounding box detection~\cite{renNIPS15fasterrcnn} (\emph{Person Size in Pixels}), as well as whether there are more than one person in the video (\emph{More than One Person}).
We look at how classification performance changes with these attributes. From Fig.~\ref{fig:person}a,c we notice that there is significant dependency on the size of the person, and that there seems to be an optimal size in the middle. This pattern was observed on multiple datasets, where models perform the best on actions with medium-sized people (48.7-78.3 pixels in size) and worse on small or large. Having multiple people in the video does not significantly affect accuracy. This suggests that the networks are not properly latching onto the person in the scene.

To investigate this further we ran ablative studies on the network. We removed the person from the scene (\emph{No Person}) and then removed everything but the person from the scene (\emph{No Background}) and evaluated the network accuracy. The removal was done by using the same Faster-RCNN bounding box detection and by setting all removed pixels to the average value. In Fig.~\ref{fig:person}b, we can see that removing the person from the test images does more damage than removing other parts of the image. Finally, we retrain the Two-stream network~\cite{simonyan2014twostream} on only the cropped image of the person (\emph{Retrain on Person}). Retraining on the cropped image yields a noticeable improvement in accuracy ($15.6\%$ to $17.9\%$ mAP). This suggests that person-focused reasoning may be beneficial to current algorithms.

\smallsec{Pose.}
Daily human activities are centered around the person. Even so, many top scoring methods get good performance without explicitly reasoning about human poses.
First, we found that the variability of a pose in each category significantly determined the accuracy on Charades ($\rho{=}0.28$). Pose variability is the average Procrustes distance~\cite{kendall1989survey} between any two poses in the category, which aligns the poses\footnote{Poses were obtained using Convolutional Pose Machines~\cite{wei2016cpm}.} with a linear transformation and euclidean distance between corresponding joints (\emph{Pose variability}). We visualize how the accuracy in a category increases as the category contains more diverse poses in Fig.~\ref{fig:person}d.
Second, pose \emph{similarity} between poses in two categories determines how much two categories are confused at test time in Charades ($\rho{=}0.39$) where we consider the same metric but across categories. 
This demonstrates that poses play a significant role in modern human activity recognition and harnessing poses is likely to lead to substantial advances.

\section{Where should we look next?}

\begin{figure*}[tb]
\centering
\includegraphics[width=1.0\linewidth]{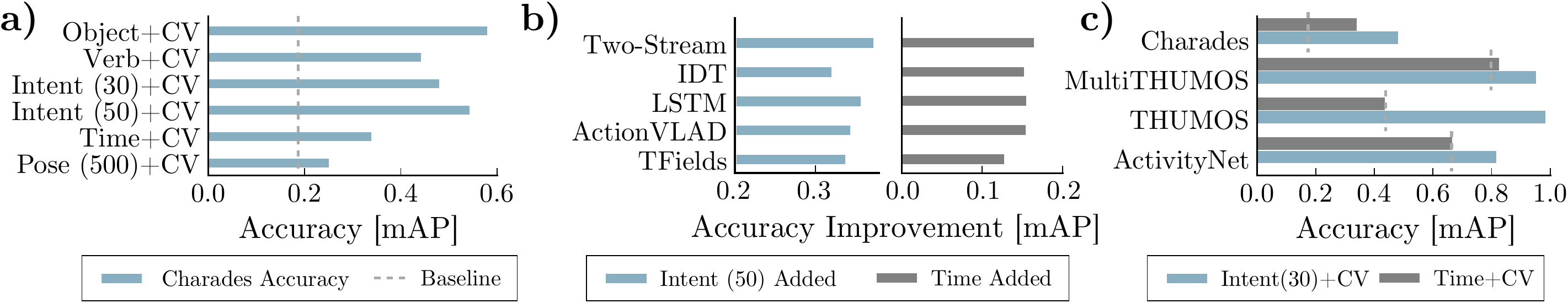}
\caption{Algorithms combined with oracles. a) Multiple types of perfect information combined with baseline on Charades. b) Two types of perfect information combined with different baselines. c) Two types of perfect information combined with a baseline on multiple datasets.}
\label{fig:priors}
\end{figure*}

Now we have analyzed state-of-the art algorithms in terms of various attributes and identified some strengths and weaknesses. To try to understand what directions look promising, we consider what aspects would help the most if solved perfectly, i.e., with an oracle. 

\smallsec{Types of oracles:} We study the effectiveness of five oracles on action recognition in Charades~\cite{sigurdsson2016hollywood}, MultiTHUMOS~\cite{yeung2015every}, THUMOS~\cite{THUMOS15}, and ActivityNet~\cite{caba2015activitynet} datasets. 

\emph{(1) Perfect Object Oracle} assumes the list of objects that the person is interacting with in the test video is given. There are 38 objects in Charades. Given this list of objects, we predict that all actions associated with these objects are present, and all other actions are absent from the video.

\emph{(2) Perfect Verb Oracle} is similar except it assumes the list of verbs that the person is executing in the test video is given. There are 33 verbs in Charades. 

\emph{(3) Perfect Temporal Oracle} assumes that for each frame of the test video, the last activity to end and the next activity to begin are given. There are 157 activities in Charades, 65 in MultiTHUMOS, 20 in THUMOS, and 200 ActivityNet. From the video annotations we learn the distribution of activities that is likely occur in this frame given this information.\footnote{We use simple first-order statistics, i.e., the probability of an action $a$ in the current frame given that action $a_{p}$ occurred before and action $a_{n}$ occurs after is $p(a|a_{p},a_{n}) = p(a|a_{p})p(a|a_{n})$.} This produces a probability distribution of actions in each frame; we max-pool over all frames to obtain predictions for the entire video. 

\emph{(4) Perfect Intent Oracle} is the trickiest. Each video contains multiple labels. When thinking about \emph{Intent} we can imagine that these labels occur together in certain ways. For example, ``put on clothes'' ``put on shoes'' ``open door'', might be associated with the intent of leaving the house.  We cluster the labels from all the videos into 30 or 50 clusters.\footnote{Each video can be thought of as a 157 dimensional vector with 0s and 1s based on what categories are in the video. We cluster these vectors with cosine distance and spectral clustering.} Each intent cluster thus corresponds to a distribution of activities. A perfect intent oracle gives which cluster the video corresponds to. Given this cluster, the distribution of activities within the cluster is used as the activity prediction.

\emph{(5) Perfect Pose Oracle} We cluster the poses in all frames into 500 clusters. Given the cluster, the distribution of activities within the cluster is used as the activity prediction. 

These oracles should be thought of as lower bounds for these types of information. That is, a method using these types of information should do at least this well on the datasets. However, it is likely that better performance could be obtained with better ways of using perfect information.

\smallsec{Comparing different oracles:} 
We start by evaluating these oracles on video classification on Charades. To do so, we design a simple video classification method using each oracle that is combined with Two-Stream (RGB)~\cite{simonyan2014twostream} on each of the datasets by multiplying their probabilities; results are presented in Fig.~\ref{fig:priors}a.
All of these oracles individually are beyond current state-of-the-art of computer vision, which suggests room for improvement in many directions.
Object understanding is more effective than temporal reasoning on its own in Charades.
This suggests object understanding is important moving forward. Getting a benefit from poses proved challenging, likely because this oracle samples individual poses throughout the video in isolation, and does not consider the motion.  
The accuracy of only 30 types of intent suggests that more research into how to understand intent is likely to yield substantial gains.\footnote{We attempted using Two-Stream to predict the 30 intent clusters in Charades (by comparing their label distributions), however this only predicted the right cluster $15.1\%$ of the time, suggesting that intent clustering should be joint with learning discriminative intent clusters. } 

\smallsec{Comparing methods in terms of oracles:} 
Next, we selected oracles on Charades to highlight differences between current methods. In Fig.~\ref{fig:priors}b we see how much is gained by combining perfect intent and time with the baselines. Adding time helps Two-Stream the most, since Two-Stream does not model temporal information beyond optical flow. Interestingly, intent helps IDT~\cite{WangIDT13} the least, which suggests the trajectory information captures high-level information missing from Two-Stream based approaches. 
To summarize, development of algorithms that combine complementary benefits of IDT, LSTM~\cite{cnnlstm} and global pooling such as ActionVLAD~\cite{Girdhar_17a_ActionVLAD} is likely to increase accuracy.

\smallsec{Comparing datasets in terms of oracles:} 
Finally, we selected two oracles to compare different datasets; results are presented in Fig.~\ref{fig:priors}c. For the datasets with fewer categories, having 30 types of intent given from the oracle, gives almost a perfect score. Although Charades and ActivityNet have 200 classes, they see improvement from only 30 types of intent. Temporal oracle helps the datasets with multiple actions per video, Charades~\cite{sigurdsson2016hollywood} and MultiTHUMOS~\cite{yeung2015every}, but not the detection oriented datasets ActivityNet~\cite{caba2015activitynet} and THUMOS~\cite{THUMOS15}. The datasets address different needs, and these results highlight that some can be useful for developing better high-level temporal modelling and others for better ways of combining predictions to detect activities.

\section{Discussion and Conclusions}

Our analysis of action recognition in videos is inspired by the diagnosis and analysis paper of Hoiem at al.~\cite{hoiem2012diagnosing} and a long line of meta-analyses that have been done in other domains: e.g., studying dataset bias in image classification~\cite{torralba2011unbiased}, analyzing sources of errors in object detection~\cite{divvala2009empirical,hoiem2012diagnosing,Khosla2012,Russakovsky2013,huh2016makes,parikh2011human}, understanding image segmentation~\cite{unnikrishnan2007toward}, and investigating specific classes of models such as CNNs~\cite{Zeiler2014} or LSTMs~\cite{Karpathy2016}. Several studies have surveyed action recognition~\cite{turaga2008machine,poppe2010survey,weinland2011survey} but to the best of our knowledge we are the first to study it in this level of depth.

We have analyzed multiple attributes of activities, several modern activity recognition algorithms, and the latest activity datasets. We demonstrated that even though human disagreement and ambiguity are an inevitable part of activity annotation, they do not present significant roadblocks to progress in activity understanding. We showed that more detailed understanding of scenes depicted in videos, at the level of individual objects and human poses, holds promise for the next iteration of algorithms. We showed that this generation of rich, multi-label, fine-grained activity benchmarks provides ample opportunities for complex joint high-level reasoning about human activities. 
We hope the community learns from our analysis, and builds upon our work.

\smallsec{Acknowledgements.}
This work was supported by ONR MURI N000141612007 and Sloan Fellowship to AG. The authors would like to thank Rohit Girdhar for supplying ActionVLAD and poses on Charades, Achal Dave for his feedback, the Deva Ramanan Lab at CMU for a helpful discussion, and the anonymous reviewers for their feedback. Finally, the authors would like to thank the authors of the algorithms and datasets surveyed in this work.

\smallsec{Appendix.}
We look at the performance among different subsets of data. Without any consideration, random chance performance would be different on different subsets. For mAP we use normalized precision $P(c)$ similar to~\cite{hoiem2012diagnosing}: $P(c){=}\frac{R(c){\cdot} N_\mathrm{pos}}{R(c) {\cdot} N_\mathrm{pos} {+}F(c) {\cdot} N_\mathrm{neg}}$,
where $R(c)$ is the recall and
$F(c)$ is the false positive rate.
N are set to the average numbers on the Charades test set.  

{\small
\bibliographystyle{ieee}
\bibliography{iccv2017gunnar.bib}
}

\end{document}